\documentclass{article}

\usepackage[utf8]{inputenc}
\usepackage[utf8]{inputenc}  
\usepackage[T1]{fontenc}     
\usepackage{lmodern}         
\usepackage{microtype}       
\usepackage[scaled]{helvet} 
\usepackage{graphicx}        
\usepackage[bookmarks=false]{hyperref}            
\usepackage{comment}         
\usepackage{booktabs}
\usepackage{multirow}
\usepackage{bm}
\usepackage{amsmath}
\usepackage[sorting=none]{biblatex} 
\title{Investigation of ensemble methods for the detection of deepfake face manipulations}

\author{
  Giatsoglou, Nikolaos\\
  CERTH \\
  \texttt{ngiatsog@iti.gr}
  \and
  Papadopoulos, Symeon\\
  CERTH \\
  \texttt{papadop@iti.gr}
  \and
  Kompatsiaris, Ioannis\\
  CERTH \\
  \texttt{ikom@iti.gr}
}
\date{\vspace{-5ex}}
\begin{document}

\maketitle
\begin{abstract}
The recent wave of AI research has enabled a new brand of synthetic media, called \textit{deepfakes}. Deepfakes have impressive photorealism, which has generated exciting new use cases but also raised serious threats to our increasingly digital world. To mitigate these threats, researchers have tried to come up with new methods for deepfake detection that are more effective than traditional forensics and heavily rely on deep AI technology. In this paper, following up on encouraging prior work for deepfake detection with attribution and ensemble techniques, we explore and compare multiple designs for ensemble detectors. The goal is to achieve robustness and good generalization ability by leveraging ensembles of models that specialize in different manipulation categories. Our results corroborate that ensembles can achieve higher accuracy than individual models when properly tuned, while the generalization ability relies on access to a large number of training data for a diverse set of known manipulations. 
\end{abstract}

\section{Introduction} 

\begin{figure}
  \centering
  \includegraphics[width=\textwidth]{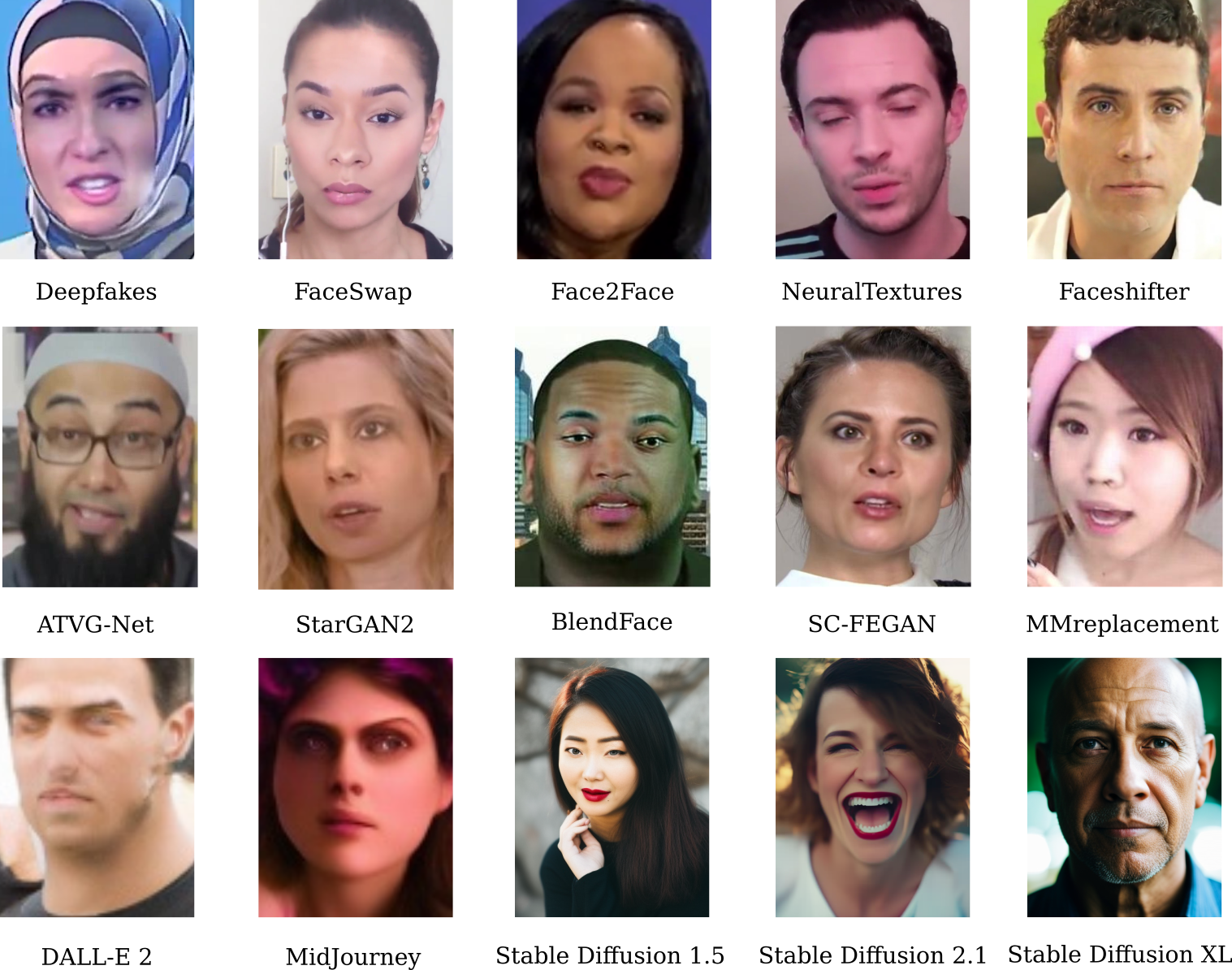}
  \caption{Examples of face forgeries created with different generation and manipulation methods. It is evident that different methods can create deepfakes of varying quality. The examples are from the FaceForensics++ \cite{rossler2019faceforensics++}, the ForgeryNet \cite{he2021forgerynet}, and the Generated-Faces-in-the-Wild \cite{borji2022generated} datasets, except for the stable diffusion images that were generated directly from the website of Stability.AI (\url{https://dreamstudio.ai}).}
  \label{fig:fakefaces}
\end{figure}

New technologies bring new challenges. This is especially true for \textit{deepfakes}, a type of synthetic media with increased photorealism, which has been recently made possible with advancements in machine learning (ML) and artificial intelligence (AI) \cite{verdoliva2020media}. On one side, deepfakes promise to revolutionize the media industry, including films and games, and to offer outlets of creativity to independent creators and Internet users. On the other side, they pose serious threats, for example, by easily generating and spreading political propaganda, revenge porn, and other types of harmful content. Since many deepfake cyberattacks rely on manipulating fake identities, research to date has been primarily focusing on human face forgeries \cite{tolosana2020deepfakes}, examples of which are shown in Fig.~\ref{fig:fakefaces}.

After acknowledging the serious threat of deepfakes, the research community has shown increased interest on appropriate detection methods. While media forensics based on conventional statistical and ML methods have been proposed, a consensus has emerged that deep neural networks are more effective at detecting the subtle artifacts of deepfakes \cite{rossler2019faceforensics++}. In particular, deep learning models can automatically learn highly discriminative features for deepfake detection without relying on laborious handcrafted feature engineering.

Deep learning techniques, however, also have limitations. First, while they achieve high detection accuracy on known manipulations, their accuracy drops considerably on unseen manipulations \cite{khodabakhsh2018fake} and is sensitive to processing operations such as compression and resizing \cite{rossler2019faceforensics++}. This is the \textit{generalization} problem of deepfake detection. Second, since forgeries are refined and become undetectable to the human eye, detection models need to rely on increasingly subtle artifacts and their decisions are hard to justify and explain. This is the \textit{explainability} problem of deepfake detection. Third, deepfake generation methods continually improve and new methods come up, so that forensics turn into an arms-race kind of problem. Finally, since deepfakes are ultimately assorted pixels like real images and videos, it is conceivable that they will be impossible to detect in the near future. Currently, there are efforts to embed provenance information in media to securely track all possible manipulations but the adoption of these technologies requires compliant capturing devices and is at a preliminary stage\footnote{\url{https://c2pa.org/}}. In the meantime, detection algorithms are worthwhile.

From the ML perspective, deepfake detection is typically treated as a classification problem, either at the video- or the image-/frame-level, supplemented by related tasks such as temporal and spatial segmentation. The deepfake class however is not clearly defined. Focusing on facial manipulations, a deepfake can be understood as a face that deviates from a representative collection of real human faces. Due to the photorealism of deepfakes, high level artifacts are becoming difficult if not impossible to detect; hence detection algorithms try to detect the subtle artifacts of the generating algorithms. Since different generation algorithms produce different artifacts, we believe that deepfake detection is best framed as an \textit{attribution} problem.

Attribution techniques have been used in the literature of generative adversarial network (GAN) images \cite{khoo2022deepfake}, where different GAN architectures are identified with specific fingerprints, and are also applicable to face forgeries. An attractive approach to attribution is through an \textit{ensemble} of AI models that specialize in different forgeries, which can readily adapt to new manipulations by incorporating new models to the ensemble. In addition, ensembles are popular in the deepfake literature because they can pool the performance of the base models and potentially generalize to unseen manipulations. Indeed, we hypothesize that a sufficiently large ensemble of manipulation methods can generalize better by detecting similarities of the unseen manipulations with the known manipulations. There are many ways, however, to build an ensemble and it is not clear which design is optimal. Due to this, the aim of our work is to experimentally investigate different designs of ensemble architectures for the detection of deepfakes.

\section{Related Work}

This section aims to give a brief overview of the deepfake detection literature in order to better situate our work within it. Section~\ref{sec:sota_tasks} describes the basic forensic tasks for deepfakes. Section~\ref{sec:sota_detection} describes the main technical approaches for detection with an emphasis on deep neural networks. Finally, Section~\ref{sec:sota_ensembles} zooms in on the ensemble techniques and their potential for deepfake detection.

\subsection{Overview of deepfake forensic tasks}
\label{sec:sota_tasks}
While traditional media forensics address operations of adding, editing, and removing objects from media, deepfakes cover a variety of manipulations of increased photorealism that are powered by AI technology. These include synthesis of completely artificial images based on a reference collection, a text input, or sketch, as well as style transfers and facial manipulations \cite{verdoliva2020media}. For the detection of these forgeries, the literature in divided in two branches: i) detection of GAN-generated images, and ii) detection of face manipulations in videos \cite{verdoliva2020media}. Focusing on face manipulations, they can be further distinguished in multiple subtypes such as \textit{face-swaps}, i.e., replacing a face with another one, \textit{face editing}, i.e., changing the characteristics of a face without replacing its identity, \textit{reenactment}, i.e., changing the expression of a face in alignment with another face, while taxonomies are continually refined. Another approach, proposed by \cite{he2021forgerynet}, is to split the face forgeries into two broad categories, \textit{identity-remained} and \textit{identity-replaced}, depending on if the identity in the tampered video is maintained. This distinction is useful because identity-remained forgeries are more challenging to detect due to the lack of blending artifacts.

The base forensic task for face manipulations is \textit{detection}, i.e., recognizing which images or video contain fake faces, which can be formulated as a binary classification problem. \cite{he2021forgerynet} has extended this task to ternary classification which tries to characterize the type of forgery as identity-remained or identity-swapped, in an effort to provide more information for forgeries. Following this line, more fine-grained tasks are possible that distinguish the exact subtype of forgery or the underlying generation algorithm. Identifying the generation algorithm is called \textit{source attribution} and is popular in the literature of GAN-generated images. This task tries to fingerprint the artifacts of GAN models and/or the noise patterns of real cameras left in authentic images \cite{khoo2022deepfake}. Attribution however is less common in the literature of face manipulation: to the authors' knowledge, \cite{jia2022model} is the only work where attribution of face-swap models is the main focus. This is partly explained because existing datasets frequently omit attribution labels to mimic a more realistic setting but the trend may be reversed with newer datasets like ForgeryNet \cite{he2021forgerynet}. In addition, \cite{jain2021improving} has produced preliminary evidence that more fine-grained discrimination of manipulations can lead to better detection and generalization to unseen manipulations.

Beyond classification problems, deepfake forensics also include segmentation tasks either in the temporal domain, i.e., finding which frames of a video are forged, or the spatial domain, i.e., finding which areas of a frame are forged. Since faces need to be extracted from images, face recognition is also relevant and used in the preprocessing step of the detection pipeline. Although this preprocessing step is frequently taken for granted, false positives can have a negative impact on the pipeline, as explored in \cite{charitidis2020investigating}. With newer deepfake datasets, algorithms are also expected to discern and keep track of multiple identities. Finally, explainability of detection, i.e., justifying the decision of detection algorithms to humans, is a widely acknowledged problem but hard to evaluate and with no standardized tasks. Currently, the literature uses heatmaps, produced by standard explainable AI (XAI) techniques, that can reveal potentially tampered areas \cite{arrieta2020explainable}. These heatmaps are evaluated qualitatively \cite{baldassarre2022quantitative} and are not straightforward to interpret, considering the photorealism of deepfakes and the subtlety of the detected artifacts.

\textbf{Our approach:} Our work focuses on detection but from an attribution perspective. Inspired by \cite{jain2021improving}, we believe that such as approach will become important for generalization as more data for known manipulations become available with newer datasets, e.g., \cite{he2021forgerynet}. We further believe that attribution can be useful for explainabilty, directing human users to  suspected types of forgery.

\begin{figure}
  \centering
  \includegraphics[width=0.7\textwidth]{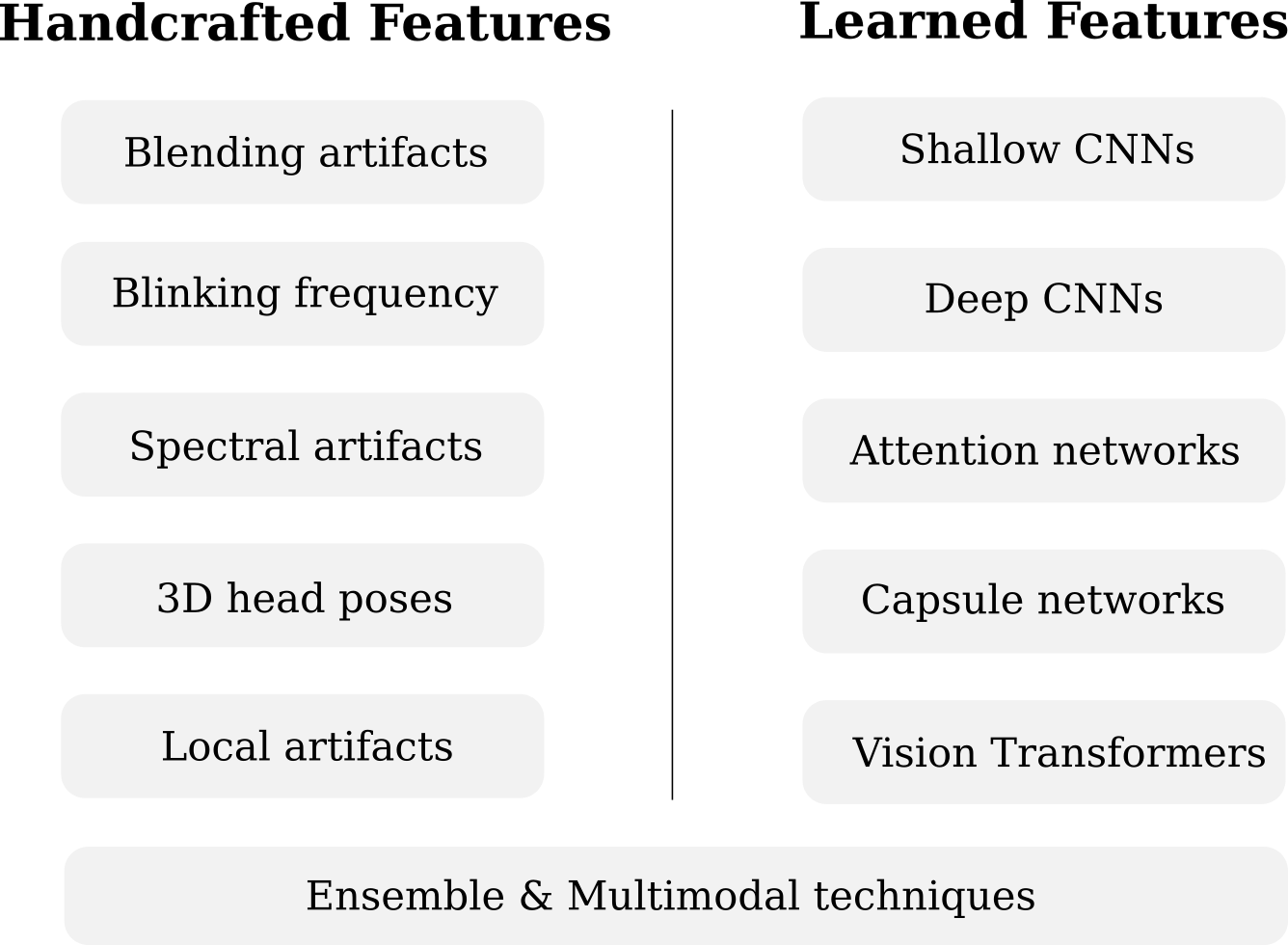}
  \caption{Representative approaches for deepfake detection with handcrafted and learned features. Notice that ensemble and multimodal techniques can be used with either type of features.}
  \label{fig:approaches}
\end{figure}

\subsection{Overview of deepfake detection methods}
\label{sec:sota_detection}
Initial approaches to deepfake detection used handcrafted features, such as scale-invariant feature transform (SIFT) vectors, image spectra, media stream descriptions, distorted facial landmarks, and biological signals \cite{masood2022deepfakes}. While these methods are easier to explain to humans, they have become less popular than neural models that automatically learn features from pixels due to their ability for optimal feature extraction, e.g., see \cite{cozzolino2017recasting}. In addition, deep neural networks are considered more effective than their shallow counterparts \cite{rana2022deepfake}, especially after the key experimental work of \cite{rossler2019faceforensics++}. Deep learning models are capable of learning both low- and high-level features of media, due to the pooling operations of subsequent layers, but are prone to overfitting: while they display > 90\% accuracy on the test set of the datasets used for training, their accuracy is significantly decreased on unseen datasets. A significant amount of work has thus focused on finding architectures that can generalize better.

From a technical perspective, deepfake detectors operate at the image-/frame- or the video-level. Frame-level detectors detect intra-frame inconsistencies and aggregate the results to the whole video. They are typically based on different flavors of convolutional neural networks (CNNs) \cite{rana2022deepfake} and, more recently, on visual transformers (ViTs) that split an image into multiple patches, e.g., \cite{coccomini2022combining}. While this approach cannot detect temporal and audiovisual inconsistencies, frame-level detector are lightweight and practical. On the other hand, video-level detectors operate on sequences of frames and, possibly, audio. They typically extract features for consecutive frames which are then passed in a temporal-aware model such as a recurrent neural network (RNN), a long short-term memory (LSTM), or a transformer \cite{rana2022deepfake}. Other approaches include 3D CNNs, capsule networks, optical flow models, and fusion techniques to incorporate audio information \cite{rana2022deepfake}. Video-level detection has more potential to detect inconsistencies but it is significantly more intensive in terms of computational resources.

\textbf{Our approach:} Our work focuses on frame-level detection due to its usefulness for general image-level detection and its use to incorporate in ensembles. Lightweight methods are also worthwhile for application in resource contrained environments, e.g., edge devices.

\subsection{Overview of ensemble techniques for deepfake detection}
\label{sec:sota_ensembles}
Ensembles of CNNs have become popular for deepfake detection, after they were found in the top solutions of the DeepFake Detection Challenge (DFDC) competition \footnote{\url{https://www.kaggle.com/competitions/deepfake-detection-challenge/leaderboard}}. Ensemble learning is an ML technique which combines multiple models so that their ensemble becomes more accurate and robust than each individual model \cite{sagi2018ensemble}. An ensemble architecture comprises the base learners and a combination layer, which can be either another learner or an aggregation function which does not require training. By combining multiple simpler models, ensembles aspire to create more accurate and robust models that generalize better. To achieve this, base models have to be reasonably accurate and diverse.

\cite{rana2020deepfakestack, kshirsagar2022deepfake} proposed ensembles of CNNs with different architectures for diversity. In particular, \cite{rana2020deepfakestack} used 7 different architectures \footnote{The architectures used were XceptionNet, ResNet101, InceptionResNetV2, MobileNet, InceptionV3, DenseNet121, and DenseNet169.} and combined their output with another CNN meta learner, while \cite{kshirsagar2022deepfake} experimented with 26 isolated CCNs and combined the ResNet models, which were found the least reliable, in order to boost their performance. In contrast, \cite{bonettini2021video} used a single CNN architecture, Efficient-B4, and diversified it with attention and different training procedures (supervised and unsupervised mode with triple loss). These works achieved high accuracy on the datasets used for training but were not tested on out-of-distribution datasets.

\cite{silva2022deepfake, silva2022deepfake} were inspired by the solutions of the DFDC challenge and experimented with the top solutions. For example, \cite{silva2022deepfake} reused the architecture of the WM team, which uses an ensemble of 3 models based on XceptionNet and EfficientNet, and incorporated attention maps for explainability and augmentation of the training dataset. \cite{trabelsi2022improving} mixed the top solutions of DFDC and experimented with different mixing strategies, in order to achieve higher accuracy on the DFDC dataset. Other related works include \cite{rao2022deepfake}, which combined existing models that detect physiological signals, and \cite{kawabe2022fake}, which ensembled models that were trained on different facial landmarks. These approaches added inductive bias in learning, which helps with their intepretability but also have the limitation of hand-crafted features. 

Finally, \cite{concas2022analysis} comprehensively studied the impact of different fusion techniques on performance, which is frequently unexplored. They used an ensemble of 6 models\footnote{The architectures used were ResNet50, XceptionNet, EfficientNet-B4 with/out attention and with supervised/unsupervised training.} and tested various parametric and non-parametric fusion methods. They found that parametric techniques work much better than non-parametric technique but their decisions are not easy to interpret and not readily extensible.

\textbf{Our approach}: From the above, we see that ensemble models are popular in the literature as a way to boost the performance of already successful models but they have not been studied consistently due to the large degrees of freedom in the ensemble design. In this work, we want to compare how different configurations of ensembles work and their impact on the generalization of deepfake detection.

\begin{figure}
  \centering
  \includegraphics[width=0.8\textwidth]{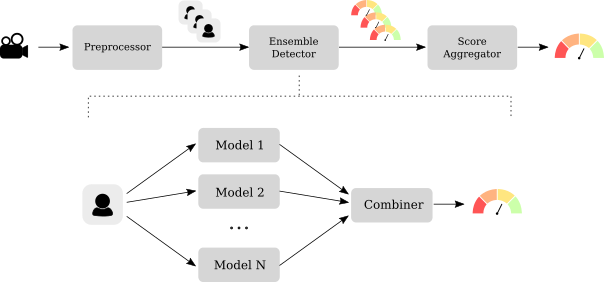}
  \caption{The architecture of our deepfake detector. The preprocessor samples frames from an input video file with a constant rate and extracts faces from them. A clustering approach that identifies faces belonging to the same identity is used in order to avoid false results in the face extraction, like in \cite{charitidis2020investigating}. The ensemble detector employs a collection of $N$ AI models to analyze the faces and a combiner to reach a robust score. Finally, the score aggregator aggregates the scores of each frame and face to arrive at a video-level score.}
  \label{fig:architecture}
\end{figure}

\section{Methodology}
\label{sec:methodology}
Our deepfake detector operates at the frame level and its architecture is shown in Fig.~\ref{fig:architecture}. The preprocessor is responsible for sampling video frames at a constant rate, recognizing and cropping the present human faces, and normalizing them to a specific size. After the faces are extracted, they are clustered to unique identities and outliers are discarded, according to the approach of \cite{charitidis2020investigating}. This step is taken in order to avoid false positives from the recognition module. Subsequently, the faces pass through the frame detector, which tries to detect the presence of forgeries, and potentially attribute the manipulation method. The resulting scores describe the confidence of the faces being fake and are eventually aggregated to derive a score for the whole video.

In the following, we focus on the design and performance of the frame detector, on which the whole pipeline depends. In particular, we first define the problem of deepfake detection formally from an ML persepctive, and then describe the ensemble design that we investigate.

\subsection{Problem setup}
\label{sec:problem_setup}
We denote a face by the triplet $(x, y, z)$ where $x$ is the face image, $y$ a multiclass attribution label, and $z$ a binary detection label. In more detail, the image $x$ is an array of dimensions $(C, H, W)$ where $C$ is the number of color channels, and $H$, $W$ are the image's height and width. Considering a selection of $K$ possible manipulations, the attribution label takes $K+1$ values where value $0$ represents a real face and the remaining values the $K$ manipulations. Finally, the detection label is binary where values $0$ and $1$ represent a real and a fake face respectively. If we know the attribution label, it is easy to derive the detection label through the following function:
\begin{equation}
z = g(y)=
    \begin{cases}
        0 & \text{if } y = 0 \\
        1 & \text{if } y > 0
    \end{cases}.
\end{equation}
Based on the above, we can define the following tasks:

\textbf{Detection task}: We seek a function $\hat{p}(z|x)$ that approximates the posterior distribution $p(z|x)$, so that:
\begin{equation}
    \hat{z} = \arg \max ~ \hat{p}(z|x)
    \label{eq:detection-task}
\end{equation}

\textbf{Attribution task}: We seek a function $\hat{p}(y|x)$ that approximates the posterior distribution $p(y|x)$, so that:
\begin{equation}
    \hat{y} = \arg \max  ~ \hat{p}(y|x)
    \label{eq:attribution-task}
\end{equation}

With a dataset offering suitable labels, it is straightforward to train single models for the detection or the attribution tasks, e.g., using the binary and the multiclass cross-entropy loss. Notice that the attribution models can be easily converted to the detection task by employing the function $g$ as $\hat{z} = g(\hat{y})$.

\subsection{Binary detection ensemble} 
For the binary detection ensemble, we group $N$ detection models, denoted by $\hat{p}_i(z|x)$, whose training has been diversified. We then soft-combine the models' outputs so that 
\begin{equation}
    \hat{p}(z|x) = \frac{\sum_{i=1}^{N} \hat{p}_i(z|x) }{N}.
\end{equation}
The final result is given by \eqref{eq:detection-task}. This ensemble cannot be used for attribution because it does not distinguish the different manipulations.

\subsection{Multiclass attribution ensemble} 
For the multiclass attribution ensemble, we group $N$ attribution models, denoted by $\hat{p}_i(y|x)$, whose training again has been diversified. We then soft-combine the models' outputs so that 
\begin{equation}
    \hat{p}(y|x) = \frac{\sum_{i=1}^{N} \hat{p}_i(y|x) }{N}.
\end{equation}
The final result is given by \eqref{eq:attribution-task} and can be used for detection by converting to binary via the function $g$.

\subsection{One-manipulation-vs-real ensemble}
For the one-manipulation-vs-real ensemble, each model specializes in one of the $K$ manipulation types and is trained as a binary classifier that discriminate between the $i$-th manipulation and the real class. Letting $s_i$ be the score of the $i$-th manipulation and $t$ a threshold, the final decision is given through max pooling as
\begin{equation}
\hat{y}=
    \begin{cases}
        \arg \max s_i & \text{if } \max \{s_i\} > t \\
        0 & \text{if } \max \{s_i\} < t 
    \end{cases},
    \label{eq:onevsreal}
\end{equation}
which can be converted to binary via the function $g$. This ensemble is very convenient because it can be easily extended with new models that specialize in different manipulations. 

\subsection{One-manipulation-vs-rest ensemble}
For the one-manipulation-vs-rest ensemble, each model specializes again in one of the $K$ manipulations but discriminates against all the remaining manipulations. Again, considering the threshold $t$ to discern the real faces, the final decision is taken via \eqref{eq:onevsreal}  The one-vs-rest ensemble is similar to the one-vs-real ensemble but delineates the boundaries of each manipulation class more clearly. The downside is that it requires the knowledge of the other manipulation classes during training, hence it is not as extensible.

\section{Results}

In this section, we present the results of our experiments. Section~\ref{sec:setup} states our experimental setup, including the used models, hyperparameters, and training procedure. The following sections then describe our results on the intra-dataset attribution task (Section~\ref{sec:intradatasetattribution}), the intra-dataset detection task (Section~\ref{sec:intradatasetdetection}), and the cross-dataset detection task (Section~\ref{sec:crossdatasetdetection}).

\subsection{Experimental setup}
\label{sec:setup}
We trained all models with the FaceForensics++ dataset which contains ground truth values about the manipulations. We used the official split of FaceForensics++\footnote{\url{https://github.com/ondyari/FaceForensics/tree/master/dataset/splits}} and preprocessed the videos as described in Section~\ref{sec:methodology} to create a dataset of face images. In particular, we sampled the videos at $1$ fps and extracted the faces with the MTCNN face detection module from the facenet-pytorch library\footnote{\url{https://pypi.org/project/facenet-pytorch/}}, which is popular in the literature. The faces are subsequently resized to $(224, 224)$ which is compatible with the model used for detection, and augmented during training for robustness similarly to the winning solution of the DFDC challenge\footnote{\url{https://github.com/selimsef/dfdc_deepfake_challenge\#augmentations}}.

For the ensemble, we used the EfficientNet-B0 CNN model \cite{tan2019efficientnet}  with pretrained weights on ImageNet for the base classifiers. The EfficientNet-B0 architecture was selected because it is performant and sufficiently lightweight in resources to construct big ensembles. The models were then trained with the PyTorch framework, using binary and multiclass cross-entropy loss, weight decay $5* 10^-4$, and the Adam optimizer with learning rate 0.001. The models were trained on the raw quality videos of the FaceForensics++ dataset for 40 epochs with an early stopping criterion of no improvement in validation accuracy after 5 consecutive epochs. To address the class imbalance, we oversampled the minority classes during training and used the balanced accuracy metric to evaluate the models.

In more details, for evaluation, we tested our models in both the \textit{intra-dataset} scenario, i.e., on the test set of the FaceForensics++ dataset, and the \textit{cross-dataset} scenario, i.e., on the Celeb-DF, DFDC preview, DFDC, and OpenForensics datasets. We evaluated both the attribution and detection task in the intra-dataset scenario, and only the detection task in the \textit{cross-dataset} scenario, since the classes of the other datasets do not match. The balanced accuracy metric is also different in the two tasks: for detection, it is defined as:
\begin{equation}
\label{eq:bacc_detection}
    BA_{det} = 0.5 ~ p(z=0|x=0) + 0.5 ~ p(z=1|x=1)
\end{equation}
while for attribution, it is defined as:
\begin{equation}
\label{eq:bacc_attribution}
    BA_{att} = 0.5 ~ p(y=0|x=0) + 0.5 ~ \frac{\sum_{i=1}^{K} p(y=i|x=i)}{K}.
\end{equation}

The intuition behind \eqref{eq:bacc_detection} and \eqref{eq:bacc_attribution} is as follows. Since we do not know the prior distribution of the real and fake faces, we give equal weights to the real and fake class, and we evenly distribute the weight of the fake class to the manipulation categories in the attribution task. In this manner, we avoid bias towards the fake classes and evaluate our models on their performance on all classes.

\begin{figure}
  \centering
  \includegraphics[width=0.8\textwidth]{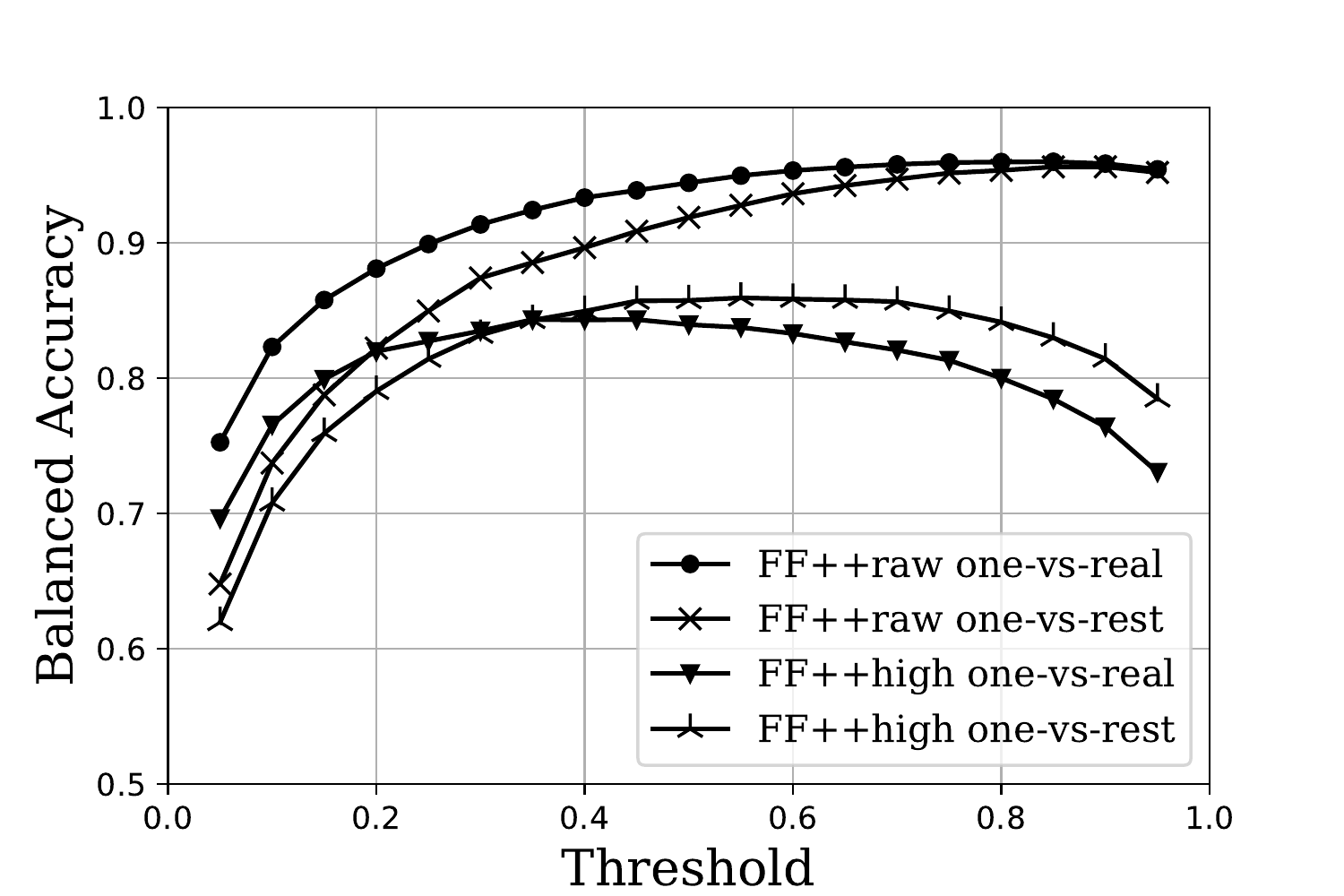}
  \caption{Balanced attribution accuracy for the raw and high quality version of the FaceForensics++ test dataset with different thresholds of the one-vs-real and one-vs-rest ensemble.}
  \label{fig:searchforthreshold}
\end{figure}

\subsection{Intra-dataset attribution task}
\label{sec:intradatasetattribution}
We first evaluate our models on the intra-dataset attribution task. Before deriving our results, we investigated appropriate thresholds $t$ for the one-vs-real and one-vs-rest ensembles. In particular, we evaluated the balanced accuracy for the attribution task on the raw and high quality version of FaceForensics++ for different values of $t$ spread evenly between $0.05$ and $0.95$ with step $0.05$. The results are shown in Fig.~\ref{fig:searchforthreshold}. We see that for the raw version of FaceForensics++, the accuracy increases monotonically with higher thresholds up to the value 0.9. This implies that the ensembles are both confident and accurate in detecting manipulations so that higher thresholds avoid false positive detections for real images. In addition, the one-vs-rest ensemble outperforms the one-vs-real ensemble for all thresholds. The behavior is different however for the high quality version of FaceForensics++ where the accuracy of both ensembles is lower and a better trade-off is achieved in the mid-range thresholds. We also see that the performance of the two ensembles is much closer and the best ensemble depends on the threshold value. For better generalization and simplicity, we selected the natural mid-range threshold $0.5$.

With these thresholds, the results of the intra-dataset attribution task are shown in Table~\ref{intra-attribution-accuracy}, which compares single attribution models with ensembles of 6 models, as is the number of classes in FaceForensics++. For the raw quality dataset, we see that the accuracy of all models is in the $>90\%$ range. In addition, the accuracy of the single models has small variation and their ensemble is better that each individual one. Finally, the one-vs-real ensemble is better than the one-vs-rest ensemble, as also seen in Fig.~\ref{fig:searchforthreshold}.

In the high quality dataset however, the behavior changes appreciably. First, there is a large variation in the accuracies of the single models, which highlights the importance of hyperparameter optimization for generalization. Indeed, considering that the single models differed only on the random parameters of the training procedure, the best single model managed to extract better features for generalization due to pure luck. Second, the accuracy of their ensemble is close to but not better than the accuracy of the best single model. This indicates that ensembles is a practical way to enhance non-optimal individual models but to ensure  best ensemble performance, parametric combination is needed. Third, the accuracies of the one-vs-real and one-vs-rest ensembles are higher and that the one-vs-rest ensemble outperforms the one-vs-real ensemble, as also seen in Fig.~\ref{fig:searchforthreshold}. This implies that the better discriminative ability of the one-vs-rest ensemble is helpful for generalization.

Finally, we see that the results on the low quality dataset are low, in some cases lower than pure chance. This shows that our model cannot distinguish manipulation artefacts when the images are heavily compressed and that a more sophisticated approach is needed for their training. Nevertheless, in the next section we evaluate if the detection models are still able to better detect general manipulation artefacts from authentic faces.

\begin{table}
\begin{center}
\resizebox{0.78\width}{!}{
\begin{tabular}{l c c c c } 
 \toprule
 & Single models & \multicolumn{3}{c}{Ensembles} \\
 \cmidrule(lr){2-2} \cmidrule(lr){3-5}
 & Multiclass & Multiclass & One vs real & One vs rest \\ [0.5ex] 
\midrule
FF++raw  & 93.35 - 96.25 & 96.45 & 94.43 & 91.88 \\
FF++high & 76.29 - 84.78 & 83.51 & 83.95 & 85.74 \\
FF++low  & 43.89 - 53.69 & 52.28 & 46.00 & 46.76 \\ 
 \bottomrule
\end{tabular}
}
\end{center}
\caption{\label{intra-attribution-accuracy} Balanced accuracy for the attribution task evaluated on the test split of the FaceForensics++ datasets.}
\end{table}

\subsection{Intra-dataset detection task}
\label{sec:intradatasetdetection}
Table~\ref{intra-detection-accuracy} shows the intra-dataset detection performance of our models. In this case, we include the multiclass attribution models that are converted to binary labels to check if the increased discrimination helps the detection task. The table includes both single multiclass models and their ensemble. In the raw quality version of the dataset, we see again that all models achieve accuracy $>90\%$. The single multiclass models can potentially achieve higher accuracy than their binary counterparts although their range of accuracies is wider. The ensembles of both types of models achieve higher performance than individual model, higher also than the one-vs-real and one-vs-rest ensembles. We note again that the one-vs-real ensemble performs better in the raw quality dataset.

As was the case in Section~\ref{sec:intradatasetattribution}, the behavior is markedly changed in the high quality dataset. Again, we note a significant variation in the accuracies of the single models, both for the binary and the multiclass cases, and that their ensembles do not achieve better performance than the individual models. Most importantly, we see that the multiclass models achieve significantly higher accuracy than the binary detectors, confirming the observations of \cite{jain2021improving}. Even more impressingly, the one-vs-real and one-vs-rest ensembles achieve the highest accuracies, suggesting the ability of specialized models to better detect forgeries than aggregate ones. 

Finally, for the low quality dataset, we again note lower accuracies, suggesting the weakness of our models to detect discriminative features under heavy compression, however, the results are improved compared with the attribution task. This implies that despite the poor attribution ability, the models are still able to distinguish some general forgeries from authentic faces. In this case, the one-vs-real achieves the best accuracy along with a single attribution model.

\begin{table}
\begin{center}
\resizebox{0.78\width}{!}{
\begin{tabular}{l c c c c c c} 
 \toprule
\multirow{2}{*}{Dataset} & \multicolumn{2}{c}{Single models} & \multicolumn{4}{c}{Ensembles} \\
 \cmidrule(lr){2-3} \cmidrule(lr){4-7}
 & Binary & Multiclass & Binary & Multiclass & One vs real & One vs rest \\ [0.5ex] 
\midrule
FF++raw  & 94.67 - 95.81 & 94.32 - 96.61 & 96.27 & 96.77 & 95.26 & 93.15 \\
FF++high & 70.71 - 80.81 & 77.02 - 86.68 & 75.50 & 83.94 & 85.71 & 87.41 \\
FF++low  & 59.60 - 63.16 & 60.11 - 65.07 & 62.58 & 62.68 & 65.36 & 63.27 \\ 
 \bottomrule
\end{tabular}
}
\end{center}
\caption{\label{intra-detection-accuracy} Balanced accuracy for the detection task evaluated on the test split of the FaceForensics++ datasets.}
\end{table}

\subsection{Cross-dataset detection task}
\label{sec:crossdatasetdetection}

We conclude with our results on the cross-dataset detection task, shown in Table~\ref{cross-detection-accuracy}. In general, we note that the accuracies are low for all datasets, suggesting the poor generalization ability of our models. Again, we see that the ensemble versions of the single models do not guarantee better performance than the individual models. Indeed, this has been a constant characteristic in all datasets except for the raw version of FaceForensics++ where our models had the best performance. This could be a prerequisite condition for the superiority of the averaging ensemble, otherwise a parametric approach is required. Additionally, the one-vs-real ensemble achieves almost always the best performance among all ensembles but, interestingly, it is comparable or even lower than the best performance of individual models. 

Despite the above, we must highlight that the perfomances are too close to random chance to extract safe conclusions. The poor generalizibility of the ensemble can be attributed to the age of the FaceForensics++ dataset. In particular, except for CelebDF, all the unseen datasets were published after FaceForensics++ and contained deepfakes of higher quality and diversity. Additionally, the manipulations of FaceForensics++ are too few to achieve good generalization. We believe that using a more recent and diverse dataset like ForgeryNet, the generalization ability of the ensembles can be significantly improved.

\begin{table}
\begin{center}
\resizebox{0.78\width}{!}{
\begin{tabular}{l c c c c c c} 
 \toprule
 & \multicolumn{2}{c}{Single models} & \multicolumn{4}{c}{Ensembles} \\
 \cmidrule(lr){2-3} \cmidrule(lr){4-7}
 & Binary & Multiclass & Binary & Multiclass & One vs real & One vs rest \\ [0.5ex] 
\midrule
CelebDF & 59.48 - 68.84 & 62.08 - 66.02 & 65.97 & 65.57 & 65.46 & 54.13 \\
DFDC preview & 55.08 - 63.29 & 54.17 - 60.39 &  59.82 & 57.28 & 64.49 & 54.15 \\
DFDC & 53.04 - 56.35 & 51.87 - 56.00 & 54.61 & 53.18 & 55.82 & 54.36  \\ 
OpenForensics & 45.50 - 57.15 & 46.22 - 55.65 & 53.32 & 51.14 & 54.89 & 49.17 \\ 
 \bottomrule
\end{tabular}
}
\end{center}
\caption{\label{cross-detection-accuracy} Balanced accuracy for the detection task on unseen datasets.}
\end{table}

\section{Conclusion}
Deepfakes are a product of the recent AI revolution, which we currently do not know how to handle. Fearing the potential damage to society and the individual, the research community has sought to find adequate solutions for the detection of deepfake media but no definitive solution has yet emerged. In this paper, following on encouraging and under-explored leads in the literature, we have investigated the potential of ensemble models to successfully detect face forgeries through attribution, aspiring to generalize on unseen manipulations. Our results have shown that, when properly tuned, ensembles can indeed achieve superior performance than individual models but a small number of manipulations is not sufficient for good generalization. In the future, we plan to enhance our solution with greater diversity of manipulations, specifically, the ForgeryNet dataset which we believe will unlock more opportunities for generalization.

\section*{Acknowledgment}
This research was supported by the EU H2020 projects TruBlo (Grant Agreement 957228) and AI4Media (Grant Agreement 951911).

\printbibliography

@article{masood2022deepfakes,
  title={Deepfakes Generation and Detection: State-of-the-art, open challenges, countermeasures, and way forward},
  author={Masood, Momina and Nawaz, Mariam and Malik, Khalid Mahmood and Javed, Ali and Irtaza, Aun and Malik, Hafiz},
  journal={Applied Intelligence},
  pages={1--53},
  year={2022},
  publisher={Springer}
}

@article{rana2022deepfake,
  title={Deepfake detection: A systematic literature review},
  author={Rana, Md Shohel and Nobi, Mohammad Nur and Murali, Beddhu and Sung, Andrew H},
  journal={IEEE Access},
  year={2022},
  publisher={IEEE}
}

@inproceedings{cozzolino2017recasting,
  title={Recasting residual-based local descriptors as convolutional neural networks: an application to image forgery detection},
  author={Cozzolino, Davide and Poggi, Giovanni and Verdoliva, Luisa},
  booktitle={Proceedings of the 5th ACM workshop on information hiding and multimedia security},
  pages={159--164},
  year={2017}
}

@inproceedings{coccomini2022combining,
  title={Combining efficientnet and vision transformers for video deepfake detection},
  author={Coccomini, Davide Alessandro and Messina, Nicola and Gennaro, Claudio and Falchi, Fabrizio},
  booktitle={Image Analysis and Processing--ICIAP 2022: 21st International Conference, Lecce, Italy, May 23--27, 2022, Proceedings, Part III},
  pages={219--229},
  year={2022},
  organization={Springer}
}

@inproceedings{rana2020deepfakestack,
  title={Deepfakestack: A deep ensemble-based learning technique for deepfake detection},
  author={Rana, Md Shohel and Sung, Andrew H},
  booktitle={2020 7th IEEE international conference on cyber security and cloud computing (CSCloud)/2020 6th IEEE international conference on edge computing and scalable cloud (EdgeCom)},
  pages={70--75},
  year={2020},
  organization={IEEE}
}

@inproceedings{bonettini2021video,
  title={Video face manipulation detection through ensemble of cnns},
  author={Bonettini, Nicolo and Cannas, Edoardo Daniele and Mandelli, Sara and Bondi, Luca and Bestagini, Paolo and Tubaro, Stefano},
  booktitle={2020 25th international conference on pattern recognition (ICPR)},
  pages={5012--5019},
  year={2021},
  organization={IEEE}
}

@article{concas2022analysis,
  title={Analysis of Score-Level Fusion Rules for Deepfake Detection},
  author={Concas, Sara and La Cava, Simone Maurizio and Orr{\`u}, Giulia and Cuccu, Carlo and Gao, Jie and Feng, Xiaoyi and Marcialis, Gian Luca and Roli, Fabio},
  journal={Applied Sciences},
  volume={12},
  number={15},
  pages={7365},
  year={2022},
  publisher={MDPI}
}

@inproceedings{kshirsagar2022deepfake,
  title={Deepfake Video Detection Methods using Deep Neural Networks},
  author={Kshirsagar, Mrunal and Suratkar, Shraddha and Kazi, Faruk},
  booktitle={2022 Third International Conference on Intelligent Computing Instrumentation and Control Technologies (ICICICT)},
  pages={27--34},
  year={2022},
  organization={IEEE}
}

@inproceedings{rao2022deepfake,
  title={Deepfake Creation and Detection using Ensemble Deep Learning Models},
  author={Rao, Sanjeev and Shelke, Nitin Arvind and Goel, Aditya and Bansal, Harshita},
  booktitle={Proceedings of the 2022 Fourteenth International Conference on Contemporary Computing},
  pages={313--319},
  year={2022}
}

@article{silva2022deepfake,
  title={Deepfake forensics analysis: An explainable hierarchical ensemble of weakly supervised models},
  author={Silva, Samuel Henrique and Bethany, Mazal and Votto, Alexis Megan and Scarff, Ian Henry and Beebe, Nicole and Najafirad, Peyman},
  journal={Forensic Science International: Synergy},
  volume={4},
  pages={100217},
  year={2022},
  publisher={Elsevier}
}

@inproceedings{trabelsi2022improving,
  title={Improving Deepfake Detection by Mixing Top Solutions of the DFDC},
  author={Trabelsi, Anis and Pic, Marc Michel and Dugelay, Jean-Luc},
  booktitle={2022 30th European Signal Processing Conference (EUSIPCO)},
  pages={643--647},
  year={2022},
  organization={IEEE}
}

@inproceedings{kawabe2022fake,
  title={Fake Image Detection Using An Ensemble of CNN Models Specialized For Individual Face Parts},
  author={Kawabe, Akihisa and Haga, Ryuto and Tomioka, Yoichi and Okuyama, Yuichi and Shin, Jungpil},
  booktitle={2022 IEEE 15th International Symposium on Embedded Multicore/Many-core Systems-on-Chip (MCSoC)},
  pages={72--77},
  year={2022},
  organization={IEEE}
}

@inproceedings{khodabakhsh2018fake,
  title={Fake face detection methods: Can they be generalized?},
  author={Khodabakhsh, Ali and Ramachandra, Raghavendra and Raja, Kiran and Wasnik, Pankaj and Busch, Christoph},
  booktitle={2018 international conference of the biometrics special interest group (BIOSIG)},
  pages={1--6},
  year={2018},
  organization={IEEE}
}

@article{verdoliva2020media,
  title={Media forensics and deepfakes: an overview},
  author={Verdoliva, Luisa},
  journal={IEEE Journal of Selected Topics in Signal Processing},
  volume={14},
  number={5},
  pages={910--932},
  year={2020},
  publisher={IEEE}
}

@article{tolosana2020deepfakes,
  title={Deepfakes and beyond: A survey of face manipulation and fake detection},
  author={Tolosana, Ruben and Vera-Rodriguez, Ruben and Fierrez, Julian and Morales, Aythami and Ortega-Garcia, Javier},
  journal={Information Fusion},
  volume={64},
  pages={131--148},
  year={2020},
  publisher={Elsevier}
}

@article{khoo2022deepfake,
  title={Deepfake attribution: On the source identification of artificially generated images},
  author={Khoo, Brandon and Phan, Rapha{\"e}l C-W and Lim, Chern-Hong},
  journal={Wiley Interdisciplinary Reviews: Data Mining and Knowledge Discovery},
  volume={12},
  number={3},
  pages={e1438},
  year={2022},
  publisher={Wiley Online Library}
}

@article{sagi2018ensemble,
  title={Ensemble learning: A survey},
  author={Sagi, Omer and Rokach, Lior},
  journal={Wiley Interdisciplinary Reviews: Data Mining and Knowledge Discovery},
  volume={8},
  number={4},
  pages={e1249},
  year={2018},
  publisher={Wiley Online Library}
}

@inproceedings{jia2022model,
  title={Model attribution of face-swap deepfake videos},
  author={Jia, Shan and Li, Xin and Lyu, Siwei},
  booktitle={2022 IEEE International Conference on Image Processing (ICIP)},
  pages={2356--2360},
  year={2022},
  organization={IEEE}
}

@inproceedings{jain2021improving,
  title={Improving generalization of deepfake detection by training for attribution},
  author={Jain, Anubhav and Korshunov, Pavel and Marcel, S{\'e}bastien},
  booktitle={2021 IEEE 23rd International Workshop on Multimedia Signal Processing (MMSP)},
  pages={1--6},
  year={2021},
  organization={IEEE}
}

@article{charitidis2020investigating,
  title={Investigating the impact of pre-processing and prediction aggregation on the deepfake detection task},
  author={Charitidis, Polychronis and Kordopatis-Zilos, Giorgos and Papadopoulos, Symeon and Kompatsiaris, Ioannis},
  journal={arXiv preprint arXiv:2006.07084},
  year={2020}
}

@article{baldassarre2022quantitative,
  title={Quantitative Metrics for Evaluating Explanations of Video DeepFake Detectors},
  author={Baldassarre, Federico and Debard, Quentin and Pontiveros, Gonzalo Fiz and Wijaya, Tri Kurniawan},
  journal={arXiv preprint arXiv:2210.03683},
  year={2022}
}

@article{arrieta2020explainable,
  title={Explainable Artificial Intelligence (XAI): Concepts, taxonomies, opportunities and challenges toward responsible AI},
  author={Arrieta, Alejandro Barredo and D{\'\i}az-Rodr{\'\i}guez, Natalia and Del Ser, Javier and Bennetot, Adrien and Tabik, Siham and Barbado, Alberto and Garc{\'\i}a, Salvador and Gil-L{\'o}pez, Sergio and Molina, Daniel and Benjamins, Richard and others},
  journal={Information fusion},
  volume={58},
  pages={82--115},
  year={2020},
  publisher={Elsevier}
}

@inproceedings{rossler2019faceforensics++,
  title={Faceforensics++: Learning to detect manipulated facial images},
  author={Rossler, Andreas and Cozzolino, Davide and Verdoliva, Luisa and Riess, Christian and Thies, Justus and Nie{\ss}ner, Matthias},
  booktitle={Proceedings of the IEEE/CVF international conference on computer vision},
  pages={1--11},
  year={2019}
}

@inproceedings{he2021forgerynet,
  title={Forgerynet: A versatile benchmark for comprehensive forgery analysis},
  author={He, Yinan and Gan, Bei and Chen, Siyu and Zhou, Yichun and Yin, Guojun and Song, Luchuan and Sheng, Lu and Shao, Jing and Liu, Ziwei},
  booktitle={Proceedings of the IEEE/CVF conference on computer vision and pattern recognition},
  pages={4360--4369},
  year={2021}
}

@inproceedings{tan2019efficientnet,
  title={Efficientnet: Rethinking model scaling for convolutional neural networks},
  author={Tan, Mingxing and Le, Quoc},
  booktitle={International conference on machine learning},
  pages={6105--6114},
  year={2019},
  organization={PMLR}
}

@article{borji2022generated,
  title={Generated faces in the wild: Quantitative comparison of stable diffusion, midjourney and dall-e 2},
  author={Borji, Ali},
  journal={arXiv preprint arXiv:2210.00586},
  year={2022}
}
\end{document}